\documentclass[letter,conference]{IEEEtran}
\usepackage{fancyhdr}
\usepackage{graphicx,times,amsmath} 
\usepackage{subfigure,comment,enumerate,morefloats,float, siunitx}
\usepackage{amssymb}
\usepackage{nicefrac}
\usepackage{xspace,multirow}

\IEEEoverridecommandlockouts
\newenvironment{tabstops}%
  {\begin{tabbing}\hspace*{0.5cm}\=\hspace*{0.5cm}\=\hspace*{0.5cm}\=\hspace*{0.5cm}\=\hspace*{0.5cm}\=\hspace*{0.5cm}\=\hspace*{0.5cm}\=\hspace*{0.5cm}\kill}%
  {\end{tabbing}}

\begin{document}
\title{Performance Measurement Under Increasing Environmental Uncertainty In The Context of Interval Type-2 Fuzzy Logic Based Robotic Sailing}
\author{\IEEEauthorblockN{Naisan Benatar, Uwe Aickelin and Jonathan M. Garibaldi {\it Member, IEEE}}
\IEEEauthorblockA{Intelligent Modelling and Analysis Research Group \\School of Computer Science\\
University of Nottingham\\
Email: [nxb,uxa,jmg]@cs.nott.ac.uk}
}
\maketitle

\author{Naisan Benatar, Uwe Aickelin\vspace{-4ex} and Jonathan M. Garibaldi {\it Member, IEEE} } 

\maketitle

\thispagestyle{fancy}
\fancyhead{}
\lhead{}
\lfoot{}
\cfoot{}
\rfoot{}
\renewcommand{\headrulewidth}{0pt}
\renewcommand{\footrulewidth}{0pt}

\begin{abstract}

Performance measurement of robotic controllers based on fuzzy logic, operating under uncertainty, is a subject area which has been somewhat ignored in the current literature.  In this paper standard measures such as RMSE are shown to be inappropriate for use under conditions where the environmental uncertainty changes significantly between experiments.  An overview of current methods which have been applied by other authors is presented, followed by a design of a more sophisticated method of comparison. This method is then applied to a robotic control problem to observe its outcome compared with a single measure.  Results show that the technique described provides a more robust method of performance comparison than less complex methods allowing better comparisons to be drawn.

\end{abstract}

\section{Introduction}

  Fuzzy logic, was initially described by Zadeh in \cite{Zadeh1965} as a generalisation of crisp set theory in which membership to a set is defined as a continuous variable between 0 and 1.  Fuzzy logic has been applied to a great number of different applications including: classification, industrial applications \cite{Mendez2009}, and many mobile robotics application, such as those by Astudillo et al \cite{Astudillo2006}. One of the main stated advantages of fuzzy logic as an approach over less sophisticated techniques such as PID (Proportional, Integral, Derivative) is that it is able to maintain higher performance in uncertain situations.

\begin{figure}[htb]
\centering
\subfigure[Type-1 Membership Function]{\label{fig:FOU0_MF}\includegraphics[scale=0.35]{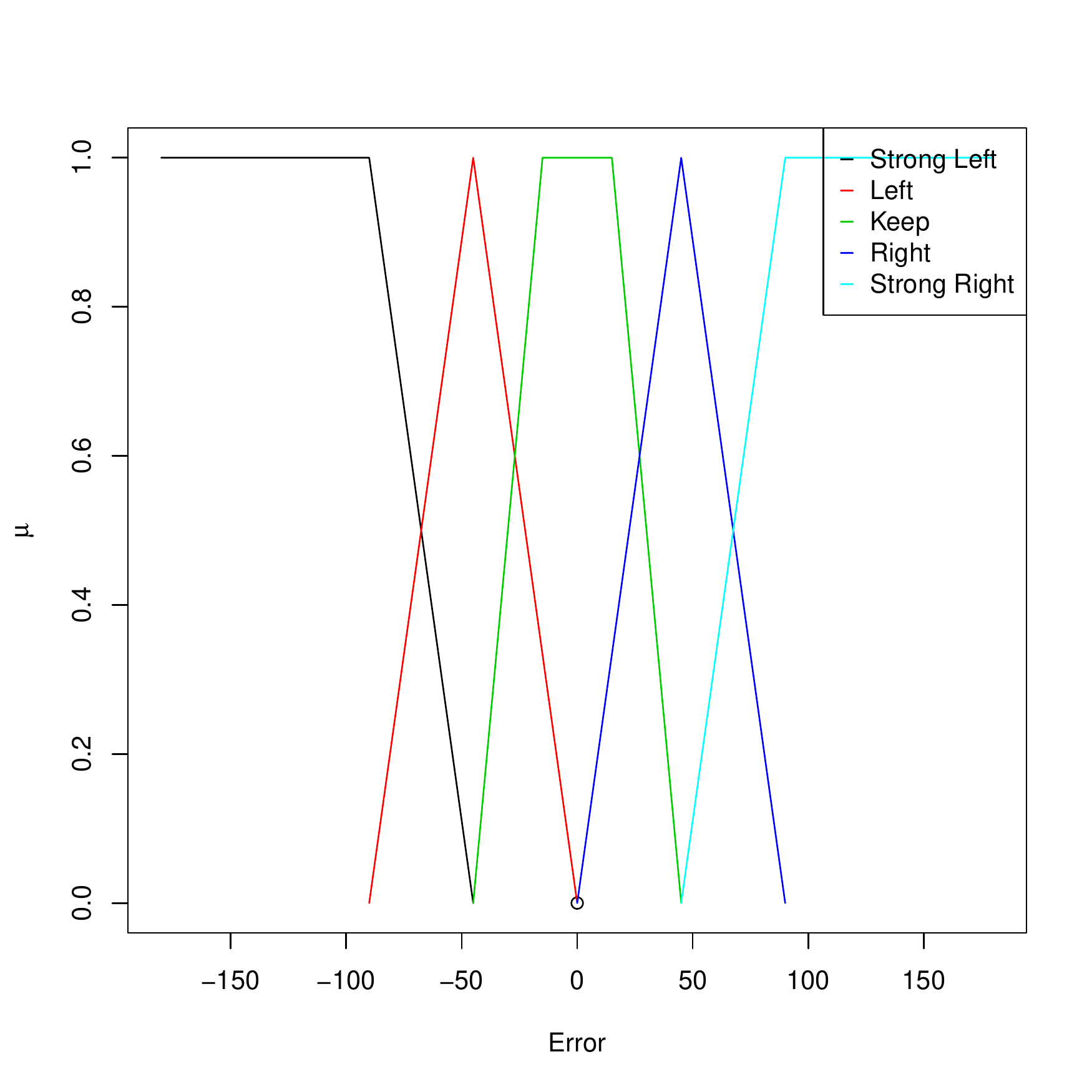}}\\
\subfigure[Interval Type-2 Membership Function]{\label{fig:FOU10_MF}\includegraphics[scale=0.35]{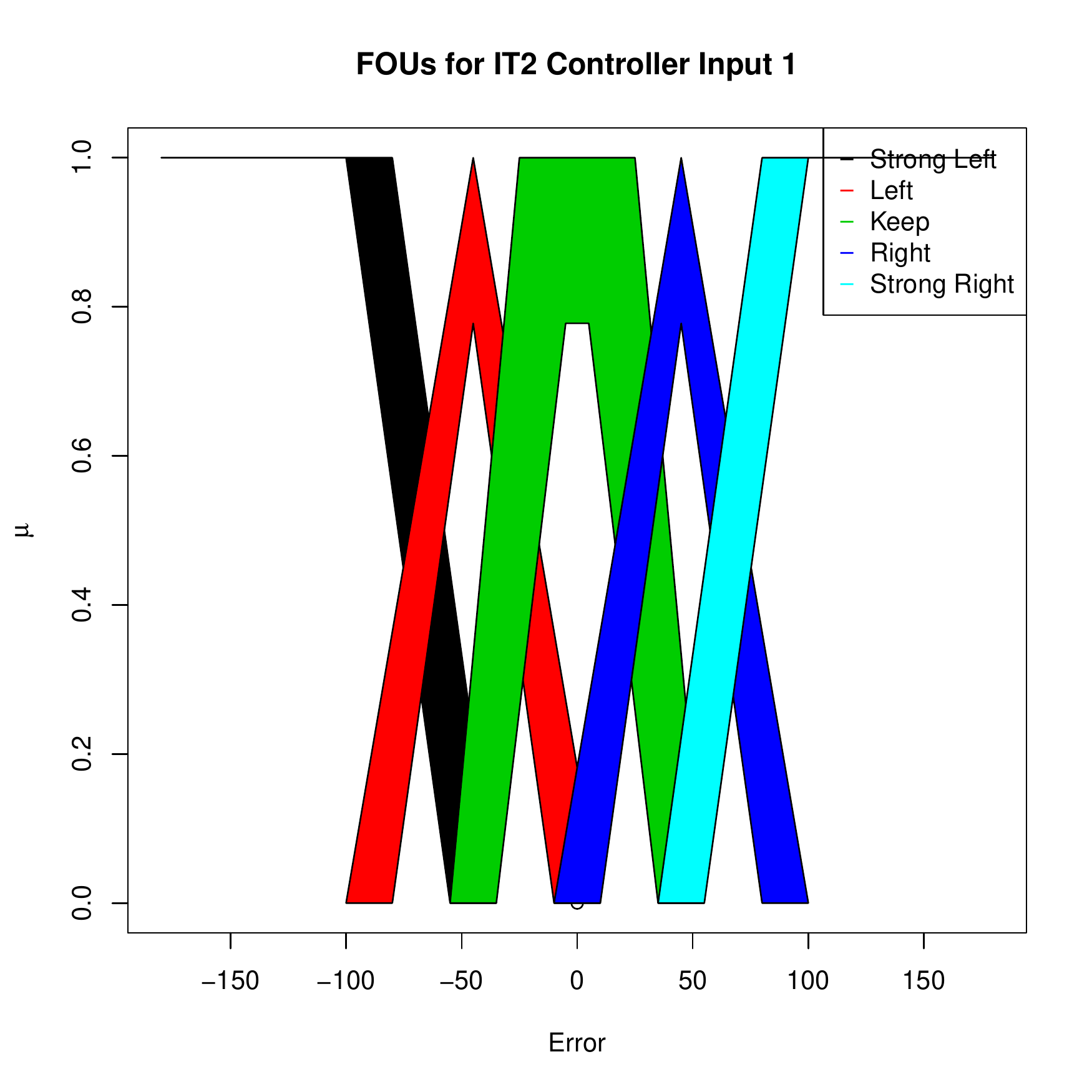}}
\caption{Example membership functions for type-1 and interval type-2 fuzzy logic systems}
\label{fig:FOUs}
\end {figure}

The initial early work on fuzzy logic in the field of robotics focused on the simplest variety of fuzzy logic, termed type-1.  This is often for reason of the low hardware requirements, which can be a considerable benefit when operating in highly resource constrained environments such as embedded systems.  In type-1 fuzzy logic the \textit{membership functions}, are simple 2-dimensional shapes, shown in Figure \ref{fig:FOU0_MF}, such as triangles which are fixed during run time.  This can cause problems when operating in highly dynamic environments where optimal placement of these functions change over time.

Type-2 fuzzy logic is considered the next step in the development of fuzzy logic systems, described by Zadeh in 1975 \cite{Zadeh1975}, in which the membership functions allow greater flexibility.  This is achieved by using either: upper and lower bounds (termed Interval Type-2 shown in \ref{fig:FOU10_MF}), or an additional dimension to create a fully 3-dimensional surface (termed general type-2). Both of these are considered improvements over type-1 fuzzy systems.  In this paper only interval type-2 is considered in which the upper and lower bounds together form an area termed the Footprint of Uncertainty (FOU).  Whether the advantage of increased flexibility of type-2 fuzzy logic offsets its additional cost in the terms of processing requirements is still very much an open question.

As the resources available to embedded systems such as robots has increased, the viability of type-2 based systems has similarly increased and has led to it becoming a more frequent subject of study in the field.  Often, the reasoning given for the selection of type-2 over type-1 fuzzy logic is the ability of type-2 fuzzy logic to better handle uncertainties present in the environment in which it operates \cite{Hagras2007}.  However without in-depth study of the effect of uncertainty upon the different controller types, it is argued that this statement is unjustified.  Additionally in many cases the metric used for comparison of different controller types and experiments is insufficient to make a strong case.

Several studies of the effect of uncertainty upon fuzzy systems have been performed such as Sepulveda et al \cite{SEPULVEDA2007} who performs a comparative work of type-1 and type-2 fuzzy systems both with and without uncertainty.   The main shortcoming with this work is that there are only two levels of uncertainty considered ``With Uncertainty'' and ``Without Uncertainty'' which we believe is insufficient to obtain a thorough understanding of its effect.  A gradual increase of the amount of uncertainty present would be allow a much stronger case to be made and further, will allow identification of the specific point at which the uncertainty present is sufficient for type-2 control to out-perform type-1. Mendel in \cite{mendel2001} and Mendel \textit{et al.} \cite{5067334a} perform similar preliminary comparisons between type-1 and type-2 fuzzy systems under differing levels of uncertainty, however they are restricted in scope.

This paper is organised as follows:  Section \ref{sec:background} provides background and an introduction into the subject of performance measurement under uncertainty.  Section \ref{sec:Metric} describes the process of the design of the comparative method which is the focus of this paper.  Section \ref{sec:SEM} discusses the experimental methodology that will be used in order to evaluate the design and utility of the final method. Section \ref{sec:Results} is where our numerical results are stated and is followed by Section \ref{sec:Discussion} in which these results are discussed in detail with reference potential shortcomings, we also draw some conclusions along with directions for future work.

\section{Background}
\label{sec:background}

	In order to show that type-2 based controllers perform better than type-1 under increased levels of uncertainty, performance measures must be carefully considered.  One of the most common metrics used for the performance measurement of robotic systems is RMSE (Root Mean Squared Error), which is the total cumulative square of the difference between the measured and desired values.  With experiments under differing levels of environmental uncertainty however, RMSE alone becomes a less meaningful measure.  This is because two separate experiments may have considerably different levels of uncertainty present but achieve the same RMSE value, this may occur for example, when conditions such as weather change between different experiments.  Using the RMSE alone as a performance measure would lead to performance being judged equal, even though one controller was operating in a more difficult, which would imply it is a better controller configuration, we feel that this should be reflected in the performance measure selected.
	
	Uncertainty is a very general term and has a large number of different usages with subtly different meanings in each. In this paper, environmental uncertainty is the main subject of consideration.  It is defined as any physical process which alters the environment during an experiment, with a prominent example being physical processes such as weather conditions.  Environmental uncertainty can cause the same set of actions to have considerably different results during an experiment and so should be taken into account when trying to compare experiments where the uncertainty levels are different.
	
	In order to take the uncertainty present into account when trying to measure performance it must be first quantified.  Because of the wide application and usages of the term uncertainty, Uncertainty Quantification (UQ) has matured into its own field.  Several different approaches have been described such as by Lee and Chen \cite{Lee2009} who focus on the how uncertainty propagates through a system.  This is in contrast to the work by Booker and Mayer \cite{Booker} who present guidelines to approach the problem of uncertainty quantification with mixed results.
	
	Fuzzy logic based controllers tend to be more commonly applied to uncertain environments than other controller types such as PID.  Specifically, a significant number of authors look to uncertainty to help design type-2 fuzzy logic systems, including Wu and Mendel in \cite{Wu2007} who present a technique in which forms of uncertainty are defined and it is investigated how these can be used to generate parameters for interval type-2 systems.  Greenfield et al \cite{Greenfield2008a} uses what is termed meta statements and meta-meta statements to quantify and describe the different types of uncertainty that can be modeled by type-1, interval type-2 and general type-2 fuzzy logic systems.  These types of approaches are high level and relatively generic, allowing them to be applied to the many different applications, which can be both a benefit and a shortcoming as in order to implement such a system it must be fitted into the specific application area.  \cite{5346145}
	
	In contrast to these more generic works above there are several approaches that are more focused on robotics applications.  Saffiotti \cite{springerlink:10.1007/3-540-49426-X_10} presents one such approach for handling uncertainty with the intention of developing robots of increasing levels of autonomy.  Lynch et al \cite{CL06} use the concept of uncertainty bounds to define parameters of an interval type-2 fuzzy logic system for the use in diesel engines --- a very specific concept which is difficult to apply to different problem spaces.
	
The problem of performance evaluation under uncertain situations has been addressed by a variety of authors, however not many publications exist in which uncertainty is considered as part of the performance measure.   As previously noted Sepulveda et al \cite{SEPULVEDA2007} discusses a study in which performance of type-1 and type-2 fuzzy logic under uncertainty is considered, however uncertainty is only considered as a binary yes or no, rather than at different levels such as `low' `medium' and `high', as done by Das et al \cite{Das2011}.  However the task under consideration by Das is very specific, involving coverage of a given area under differing levels of sensor uncertainty which is not strictly environmental in nature making it less applicable to our problem.

	From the discussion of the existing literature we have identified a gap in the field of performance measurement under increasing levels of environmental uncertainty.  This has led to the decision to develop a method for the performing comparisons of experiments run under differing levels of environmental uncertainty. The eventual aim is to be able to answer questions such as ``At what level of uncertainty does a type-2 fuzzy controller out perform a type-1 controller?'' with an answer such as ``At a Environmental uncertainty level of 0.75 a type-2 controller will outperform a type-1 controller with a confidence of 0.9''.<>

\section{Metric Design}
\label{sec:Metric}

	In order to develop a performance measure that incorporates uncertainty we must first analyse the uncertainty present in the application under consideration.  Based on previous work \cite{Benatar2012} it has been shown that the FLOATS (Fuzzy Logic Operated AuTonomous Sailing Boat) platform provides a good base for which such development can be built on.  The application is, as the name suggests, a simulated autonomous boat that is designed to move between defined way points using the wind in the same way a human sailor would --- by controlling the rudder and sail positions.
	
	The simulator used as part of the FLOATs platform provides one main source of uncertainty --- a wind source which is defined by four parameters, the maximum and minimum wind speed (specified in m/s) and the maximum and minimum wind directions (specified in degrees with 0 indicating due north).  The wind speed and direction will be considered separate sources in this context as in this simulation they are controlled independently.  They will be the basis of the Uncertainty measure that will be used to weight the overall performance of a given experiment.  While real world situations will involve considerably more sources with potentially greater magnitude and complexity, for development purposes the simulated process is considered sufficient, especially as it gives fine grain control over each uncertainty source, which is a very desirable property in this developmental phase.  From a contextual point of view, a value of 0 will indicate that there is no uncertainty i.e. the environment is deterministic and increasing value will indicate a larger amount of uncertainty, a maximum value has been defined as 1, forced by normalisation.
	
	The environmental sources of uncertainty present during the experiments will be quantified into two variables termed ``Uncertainty Measure'' and ``Base Difficulty''.  The total uncertainty when the two sources described above are considered, can be obtained by calculation of the product of their standard deviations.  The ``Base Difficulty'' is intended to quantify how difficult a given environment is if no uncertainty was present. This can be derived in this situation by looking at the mean wind direction and speeds, and is best illustrated using the wind direction:  If the wind did not change and was blowing directly into the boat, it would not be possible to move towards the target causing the ``Base Difficulty'' to be high. Conversely, if the wind was blowing from directly behind the boat it would be easy to move towards the goal, meaning ``Base Difficulty'' would be correspondingly low.  We have observed during preliminary tests the mean wind speed within limits does not make a significant effect on the boat performance which has led to use applying the filtering method shown in Figure \ref{fig:WindSpeedValA}.
	
	\begin{figure}
\begin{center}
\fbox{\begin{minipage}{10cm}
{\tt
\begin{tabstops}
		if ( Wind Speed $==$ 0 )\\
			Wind Speed Value = 0 \\
		else\\
			\>if ( Wind Speed $>$ 14 )\\
			\>\>	Wind Speed Value = 1\\
			\>else\\
			\>\>	Wind Speed Value = 0.5\\
		endif
\end{tabstops}}
\end{minipage}}
\caption{Method for selecting Wind Speed Value}
\label{fig:WindSpeedValA}
\end{center}
\end{figure} 
	
	One of the most commonly used performance measures for this variety of application as previously stated is RMSE (Root Mean Square Error), which measures how far from an 'optimal course' a particular experiment lies.  As this is the goal of the experiment being conducted here, the RMSE will serve as a basis for what is termed ``Absolute Performance'', which aims to provide a raw, unweighted measure of performance.  The standard way of looking at RMSE is that a lower value (closer to 0) indicates better performance and this ordering will be maintained for the Absolute Performance value with a value of 0 indicating best possible performance.  
	
\begin {figure*}[htb]
\centering
\includegraphics [scale=0.15]{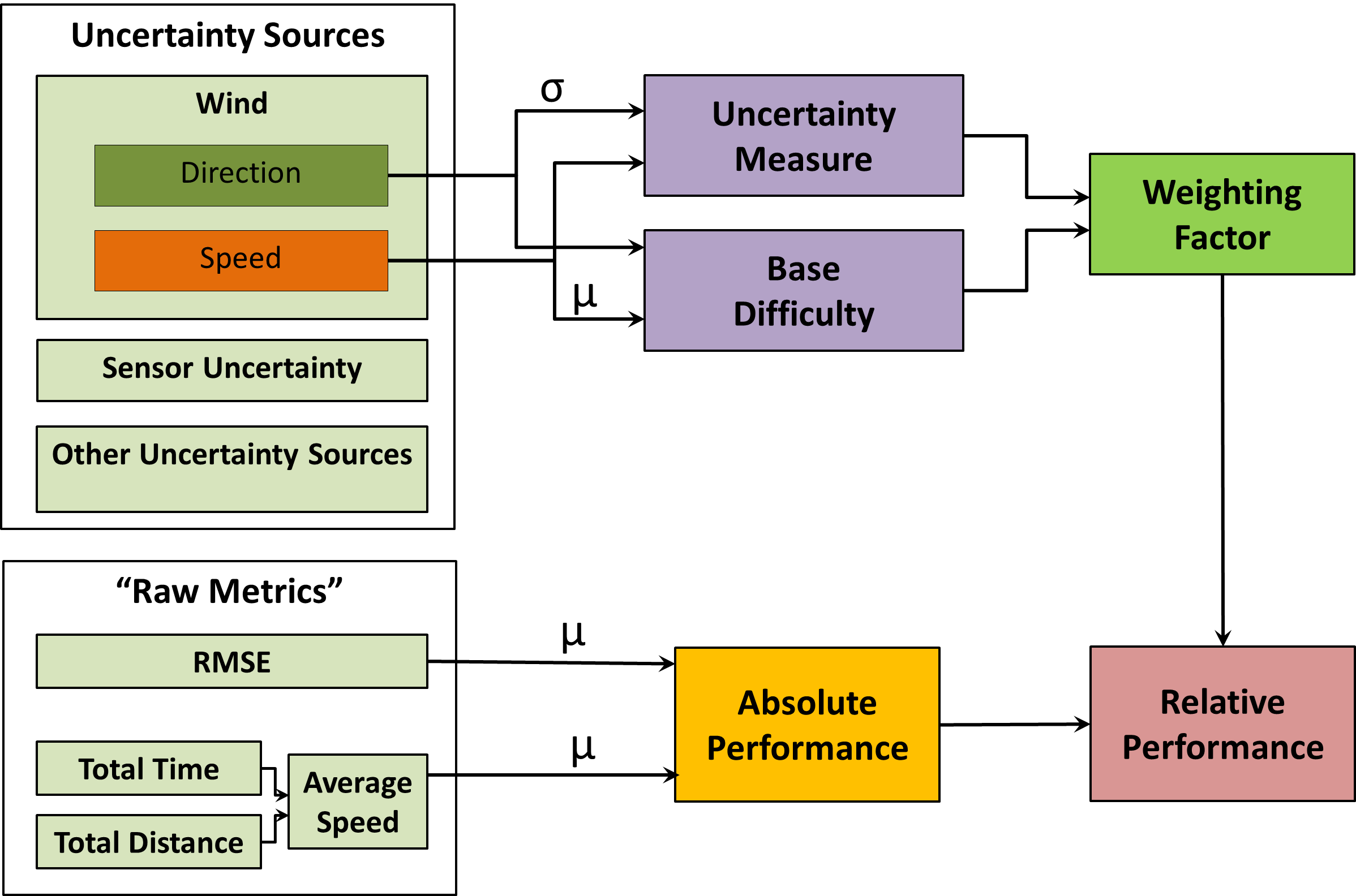}
\caption{Data Flow for Metric Design Process.  The left-most values in green indicate values that can be calculated from basic data collected by the boat.  Purple and orange objects indicate values which combine aspects (either mean or standard deviations) of the basic values.  This, then flows into the final measure, `Relative Performance' which will be used for comparisons between numerous different experimental set-ups.}
\label{fig:MetricDesign}
\end {figure*}

	Figure \ref{fig:MetricDesign} shows the flow of information used in the development of the performance metric from the source through to final Relative Performance value.  This leads us to being able to define Equations \ref{eq:UncertaintyMetric}, \ref{eq:AbsPerf}, \ref{eq:BaseDifficulty} and \ref{eq:RelPerf} which illustrates each input into the system and how they are fed into the final measure, Relative Performance.  The method for combining each input into of the formulae must also be defined before this technique can be applied to data.
	
	In order to keep the formulae as simple as possible all functions within each of the shown equations will be defined as the product its inputs.  The only exception to this rule will be the relative performance equation, in which the absolute performance will be divided by the product of the uncertainty measure and base difficulty meaning the final relative performance will be the absolute performance weighted by the uncertainty present as intended and which gives the final relative performance formulae shown in Equation \ref{eq:FinalEq}.  It should be noted that we maintain the ordering of RMSE i.e. a lower value indicates better overall performance.  

\begin{figure}[htb]
\centering
\subfigure[Change of standard deviation of wind speed]{\label{fig:Exp1_Speed}\includegraphics[scale=0.15]{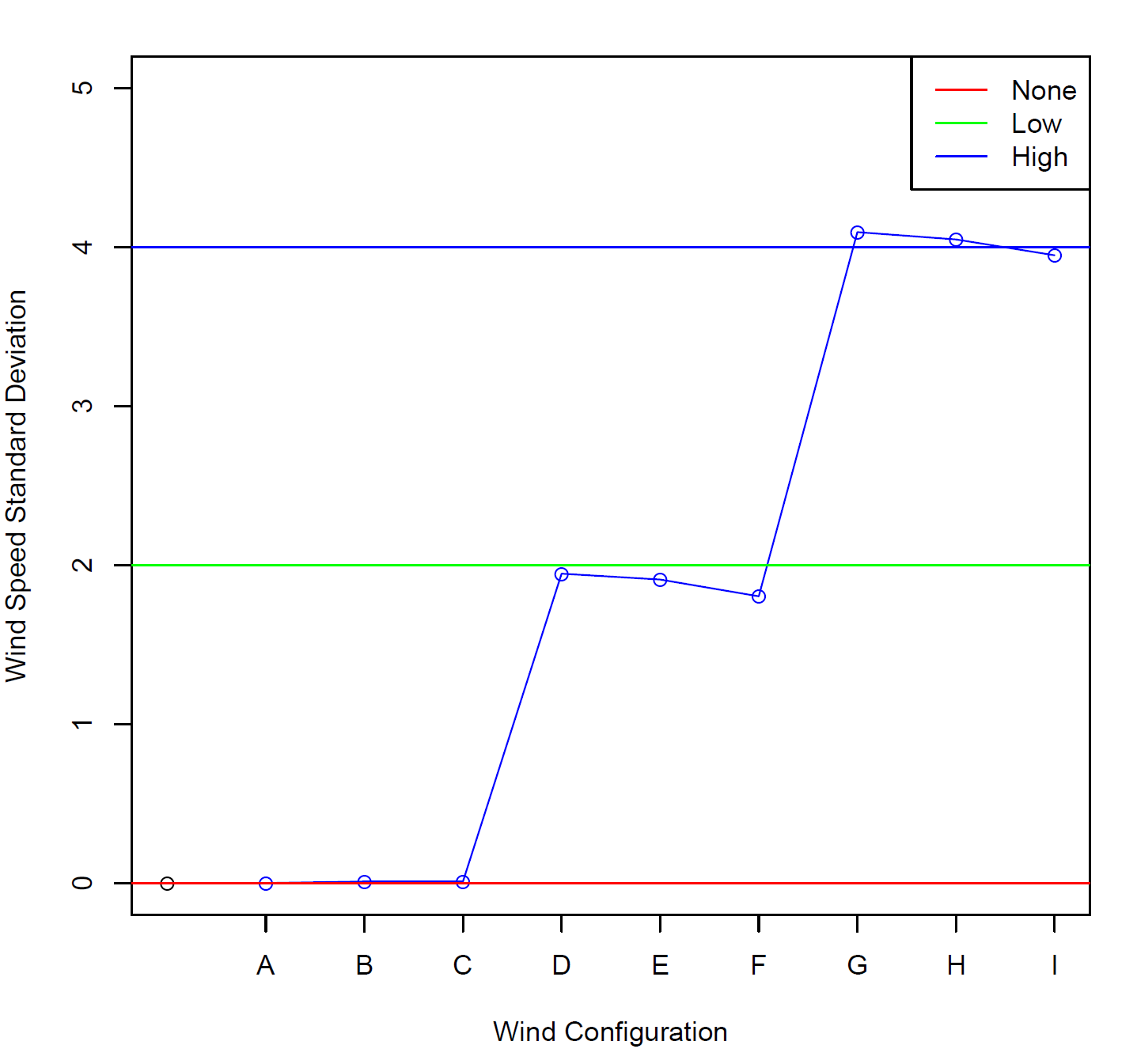}}\\
\subfigure[Change of standard deviation of wind direction]{\label{fig:Exp1_Direction}\includegraphics[scale=0.15]{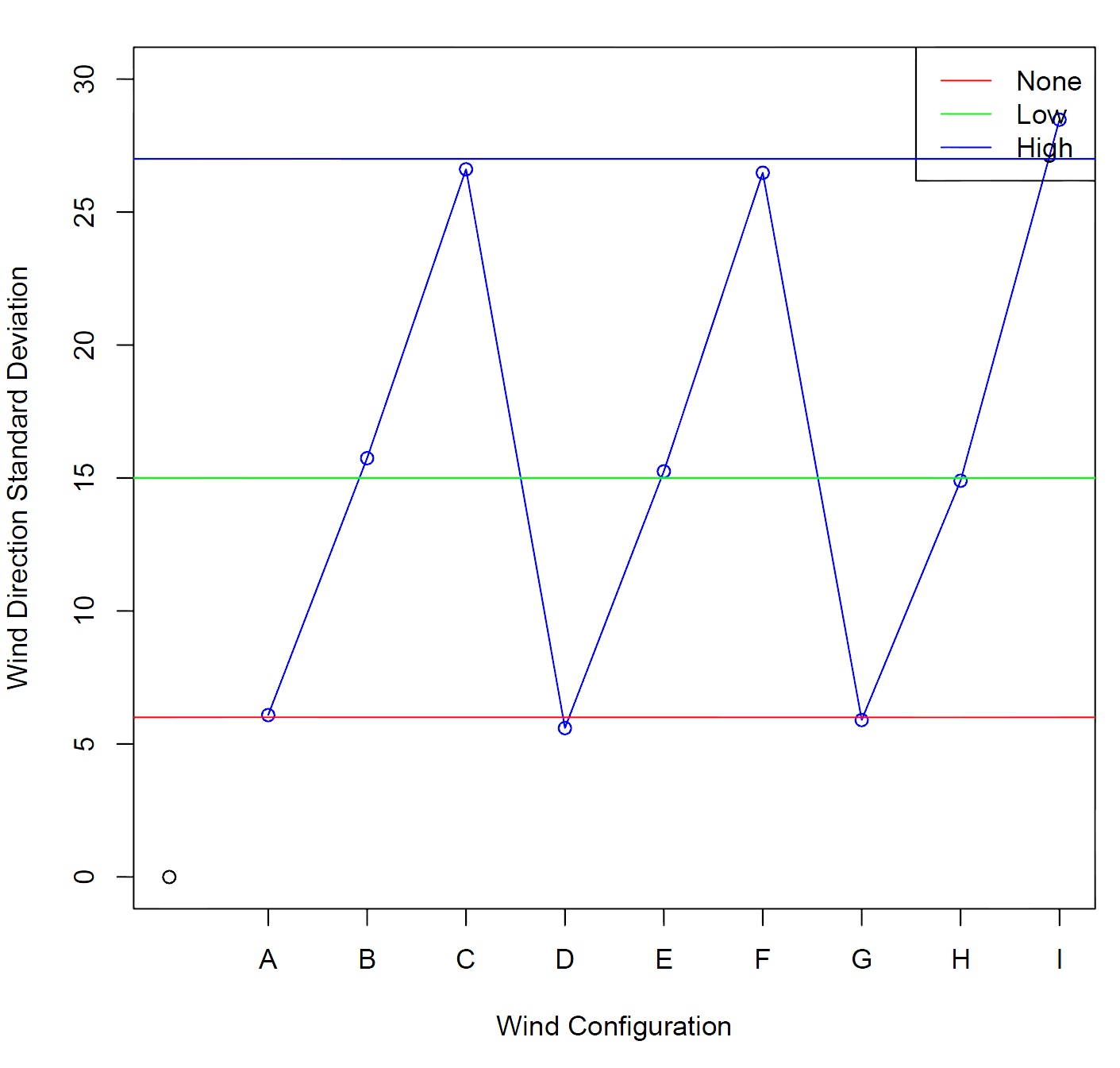}}
\caption{Standard deviations of wind speed and directions as wind configurations are changed}
\label{fig:UncertaintyMetricInputs}
\end {figure}

Figure \ref{fig:UncertaintyMetricInputs} shows the standard deviations of the Wind Speed (Figure \ref{fig:Exp1_Speed}) and Wind Direction (Figure \ref{fig:Exp1_Direction}) in order to ensure that the inputs used in the calculation of the uncertainty measure are correct.  It can be observed that in both figures three different levels of uncertainty are obvious, indicated by overdrawn red (low), green (medium), and blue (high) lines.  In the case of the wind direction, the saw-tooth pattern can be explained by looking at Table \ref{tab:Configs} and reading it row by row, giving configurations `A', `D' and `G' one value indicating no uncertainty, configurations `B', `E', `F' the next value and finally configurations `C', `F' and `I' the highest level of uncertainty.  In a similar manner, the pattern present in the wind speed graph can be explained by reading the table in a column by column manner, meaning configurations `A', `B' and `C' would have the similar values as would configurations `D', `E' and `F' and so forth.

\begin{equation}
\label{eq:RMSE}
RMSE_{Boat} = \sqrt{\frac{(Bearing_{Desired} - Bearing_{Actual})^2}{n}}
\end{equation}

\begin{equation}
\label{eq:AbsPerf}
{Perf}_{Absolute} = f(RMSE_{Boat})
\end{equation}

\begin{equation}
\label{eq:UncertaintyMetric}
Uncertainty_{Measure}= h(\sigma(Wind Direction),\sigma(Wind Speed))
\end{equation}

\begin{equation}
\label{eq:BaseDifficulty}
Base Difficulty = g(\mu(Wind Direction),\mu(Wind Speed))
\end{equation}

\begin{equation}
\label{eq:RelPerf}
Perf_{Relative} = i(Uncertainty Measure, Perf_{Absolute})
\end{equation}

\begin{equation}
\label{eq:FinalEq}
Perf_{Relative} = \frac{Performance_{Absolute}} {Uncertainty Measure * Base Difficulty}
\end{equation}

	It is anticipated that the newly deigned metric will, as wind configuration increases from `A' to `I', and therefore uncertainty is increased, exhibit a downwards curve as the denominator shown in Equation \ref{eq:FinalEq} will increase.  Previous studies have shown that RMSE does not follow a fixed pattern as the uncertainty is varied and it is hoped the relative performance will show a more regular output.
	
\section{Experimental Methodology}
\label{sec:SEM}

	As previously discussed, the FLOATS platform will be used as a test bed for generating data that can be used in order to evaluate the use of this technique.  Previous work \cite{Benatar2011} has shown that FLOATs is viable for this sort of data, but that using RMSE as a performance metric does not give a strong correlation or any obvious patterns over all the different combinations that have been tested.  It is anticipated that the results found using the Relative Performance will enable better comparisons and stronger conviction as to the conclusions made than when RMSE alone is used.

\subsection{Experimental Design}
\label{sec:Methodology}

	In this paper the type-2 footprints of uncertainty used in the interval type-2 controllers will be derived from the type-1 by introducing a horizontal movement with the degree of movement altered to give six different widths and therefore FOU sizes: 0, 5, 10, 15, 20 and 25.  This is done in order ensure that the relative performance metric is robust and operates with different FOU values, FOU sizes 0 and 10 are show in Figures \ref{fig:FOU0_MF} and \ref{fig:FOU10_MF} respectively.	
	
	The controller set-up is the same that has been used in previous work by the authors (\cite{Benatar2011}and \cite{Benatar2012}.  Two input variables are used (error and delta error) each with five associated fuzzy sets, which leads to 25 rules being used.  The output fuzzy sets are five singletons with the exact values shown in the previous work.

	Uncertainty in these experiments is introduced using a simulated wind process which is controlled from within the simulator and which can be assigned limits in the form of a parameter file.  Table \ref{tab:Windspeed} outlines the different values of wind speed and direction that will be used in this experiment while Table \ref{tab:Configs} enumerates how these are combined to give environmental uncertainty configurations `A' through `I'.  This gradual change should cause the uncertainty measure to start at a value of 0 for configuration `A' to the highest value on configuration `I' and this fine grained control should allow and observation as to what the effect of uncertainty is upon each experiment.  In this case the uncertainty score is only used to allow ordering of the different configurations.
	
\begin{table}[tb]
\caption{Wind Speed and Direction Upper and Lower Values and Uncertainty Score.  Speed measured in m/s and direction in degrees}
\begin{tabular}{lrrr}
\hline
Direction & \multicolumn{1}{l}{Uncertainty Score} & \multicolumn{1}{l}{Lower Limit} & \multicolumn{1}{l}{Upper Limit} \\ \hline
None & 0 & 180 & 180 \\ 
Low & 1 & 160 & 200 \\ 
High & 2 & 140 & 220 \\ 
&&& \\ 
\hline\hline
Speed & \multicolumn{1}{l}{Uncertainty Score} & \multicolumn{1}{l}{Lower Limit} & \multicolumn{1}{l}{Upper Limit} \\ 
None & 0 & 7 & 7 \\ 
Low & 1 & 4 & 10 \\ 
High & 2 & 1 & 13 \\ 
\hline
\label{tab:Windspeed}
\end{tabular}
\end{table}

\begin{table}[h]
\begin{center}
\caption{Wind Configuration Definitions}
\begin{tabular}{ll|llll}
\hline
Wind Speed $\rightarrow$ & & \textbf{None} & \textbf{Low} & \textbf{High}  \\ \hline
\multicolumn{ 1}{l}{Direction $\downarrow$} & \textbf{None} & A & D & G \\ 
\multicolumn{ 1}{l}{} & \textbf{Low} & B & E & H \\ 
\multicolumn{ 1}{l}{} & \textbf{High} & C & F & I \\ 
\hline
\end{tabular}
\label{tab:Configs}
\end{center}
\end{table}

\begin{figure}[htb]
\centering
\includegraphics[scale=0.50]{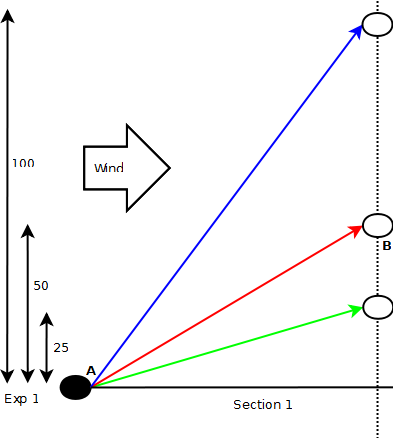}
\caption{Each coloured line represents a course layout under test.  The white circles represent end points and the black circle the start point. The angles required for the first turn are 5.71$^{\circ}$, 11.42$^{\circ}$ and 21.84$^{\circ}$ for 25, 50 and 100 meters vertical movements respectively.  Not to scale.}
\label{fig:Course}

\end{figure}

Figure \ref{fig:Course} shows the three different course layouts that will be used in these experiments which vary by the angle required to complete the course defined as either 25, 50 or 100 metres.  This equates to turns of 5.71$^{\circ}$, 11.42$^{\circ}$ and 21.84$^{\circ}$ for 25, 50 and 100 meters vertical movements respectively.  It is anticipated that a larger turn angle would be more difficult to perform meaning that we would expect to see performance get worse as it is increased.

	Every combination of course and wind configurations will be tested with each different FOU size with the mean of 30 repeats being used for each calculation.  We will start with no noise (configuration `A') and move towards the most uncertain environment (Configuration `I'). Every four seconds a wind change will be triggered by the simulator using a Gaussian random number generator to change the values of the wind speed and direction.  
	
\section{Results}
\label{sec:Results} 

Table \ref{Tab:FOU10Metrics} shows raw data from an experiment in which a controller with an FOU size of 10 controls the boat on a course in which the vertical movement is 50m and which each wind configuration is used.  We can see that the uncertainty measure matches our hypothesis and increases in a regular manner with Configuration `I' showing the largest, and configuration `A' the smallest values.  The absolute performance is essentially just the raw RMSE value and therefore follows the same lines in that over all the different configurations combinations there are in general few patterns which hold true at every point.  In the final column, the relative performance is shown and it can clearly be seen that it decreases in a regular pattern as the wind configuration transitions from  `A' to `I', indicating that when uncertainty is taken into account the relative performance actually improves.

\begin{table}[htbp]
\caption{Calculated variable values for all wind configurations for experiments with a horizontal movement of 50m and a FOU size of 0}
\begin{tabular}{lrrrr}
\hline
Wind & Uncertainty & $Perf_{Absolute}$ & Base & $Perf_{Relative}$ 	 \\ 
Config. & Measure & & Difficulty & $\times 10^2$ \\ \hline
A & 14.05 & 6.54 & 33.38 & 139.54 \\ 
B & 23.98 & 7.67 & 30.48 & 104.81 \\ 
C & 32.37 & 6.83 & 28.41 & 74.22 \\ 
D & 42.36 & 6.59 & 34.40 & 45.22 \\ 
E & 61.14 & 6.14 & 30.34 & 33.09 \\ 
F & 88.40 & 5.79 & 28.08 & 23.31 \\ 
G & 81.60 & 7.14 & 35.42 & 24.69 \\ 
H & 120.76 & 7.32 & 32.70 & 18.53 \\ 
I & 152.34 & 6.20 & 28.89 & 14.09 \\ \hline
 \end{tabular}
\label{Tab:FOU0Metrics}
\end{table}

\begin{table}[htbp]
\caption{Calculated variable values for all wind configurations for experiments with a horizontal movement of 50m and a FOU size of 10}
\begin{tabular}{lrrrr}
\hline
Wind & Uncertainty & $Perf_{Absolute}$ & Base & $Perf_{Relative}$ 	 \\ 
Config. & Measure & & Difficulty & $\times 10^2$ \\ \hline
A & 14.21 & 6.76 & 33.68 & 141.23 \\ 
B & 24.29 & 7.53 & 30.60 & 101.33 \\ 
C & 33.85 & 7.26 & 29.07 & 73.76 \\ 
D & 43.19 & 6.70 & 34.44 & 45.03 \\ 
E & 67.21 & 6.69 & 31.17 & 31.93 \\ 
F & 94.31 & 6.11 & 28.61 & 22.65 \\ 
G & 85.08 & 7.46 & 35.90 & 24.42 \\ 
H & 125.42 & 7.39 & 33.00 & 17.85 \\ 
I & 161.64 & 6.36 & 29.31 & 13.43 \\ \hline
 \end{tabular}
\label{Tab:FOU10Metrics}
\end{table}

\begin{table}[htbp]
\caption{Calculated variable values for all wind configurations for experiments with a horizontal movement of 50m and a FOU size of 20}
\begin{tabular}{lrrrr}
\hline
Wind & Uncertainty & $Perf_{Absolute}$ & Base & $Perf_{Relative}$ 	 \\ 
Config. & Measure & & Difficulty & $\times 10^2$ \\ \hline
A & 13.51 & 6.48 & 33.73 & 142.08 \\ 
B & 24.61 & 7.40 & 30.71 & 97.97 \\ 
C & 34.39 & 7.06 & 29.39 & 69.87 \\ 
D & 41.03 & 6.33 & 34.27 & 45.03 \\ 
E & 70.75 & 6.88 & 31.80 & 30.58 \\ 
F & 96.74 & 6.04 & 28.95 & 21.55 \\ 
G & 83.85 & 7.28 & 36.02 & 24.09 \\ 
H & 123.39 & 6.77 & 32.97 & 16.65 \\ 
I & 165.83 & 6.25 & 29.63 & 12.71 \\ \hline
 \end{tabular}
\label{Tab:FOU20Metrics}
\end{table}

\begin{figure*}[htb]

\subfigure[RMSE for 25 Turn]{\label{fig:RMSE_25}\includegraphics[scale=0.3]{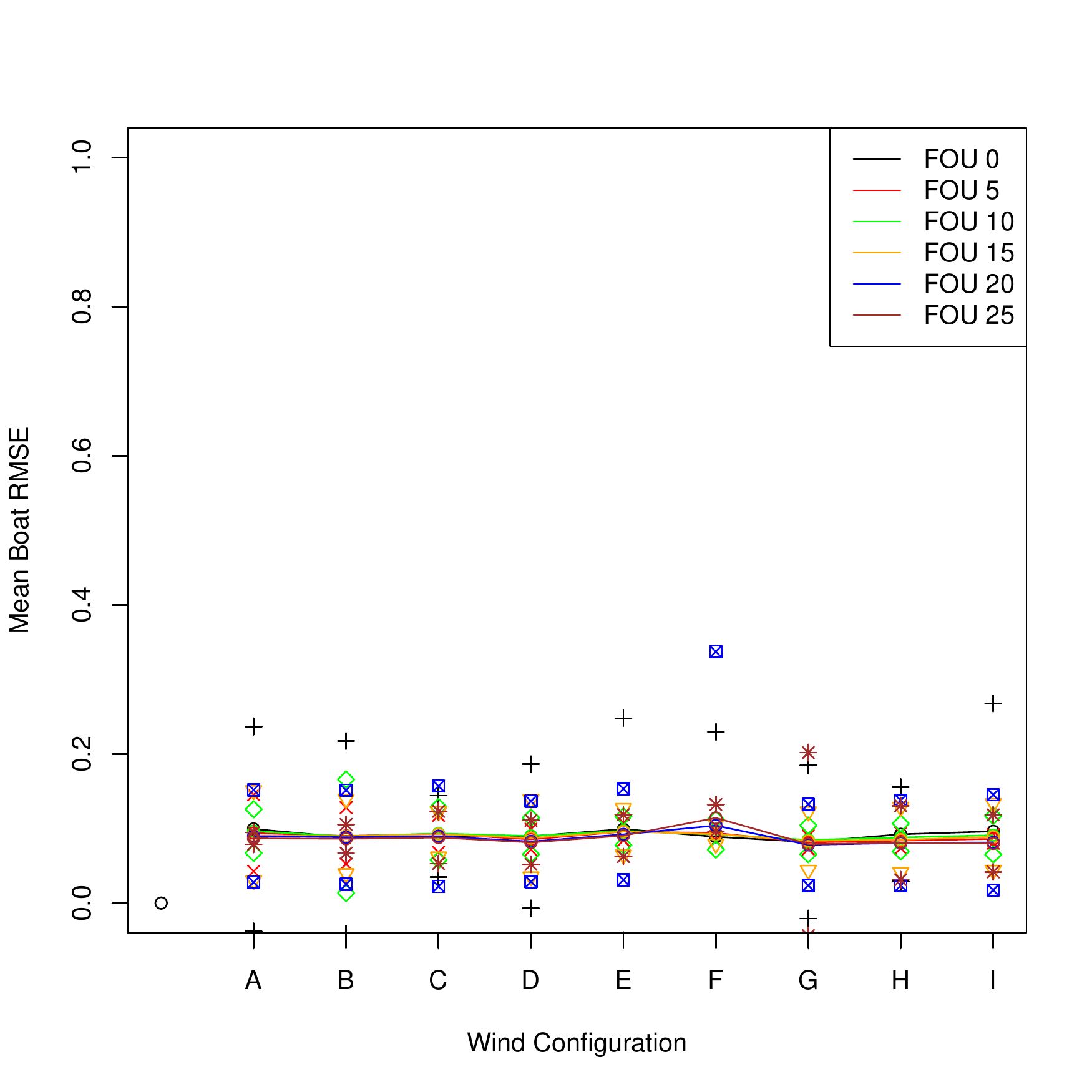}}
\subfigure[RMSE for 50 Turn]{\label{fig:RMSE_50}\includegraphics[scale=0.3]{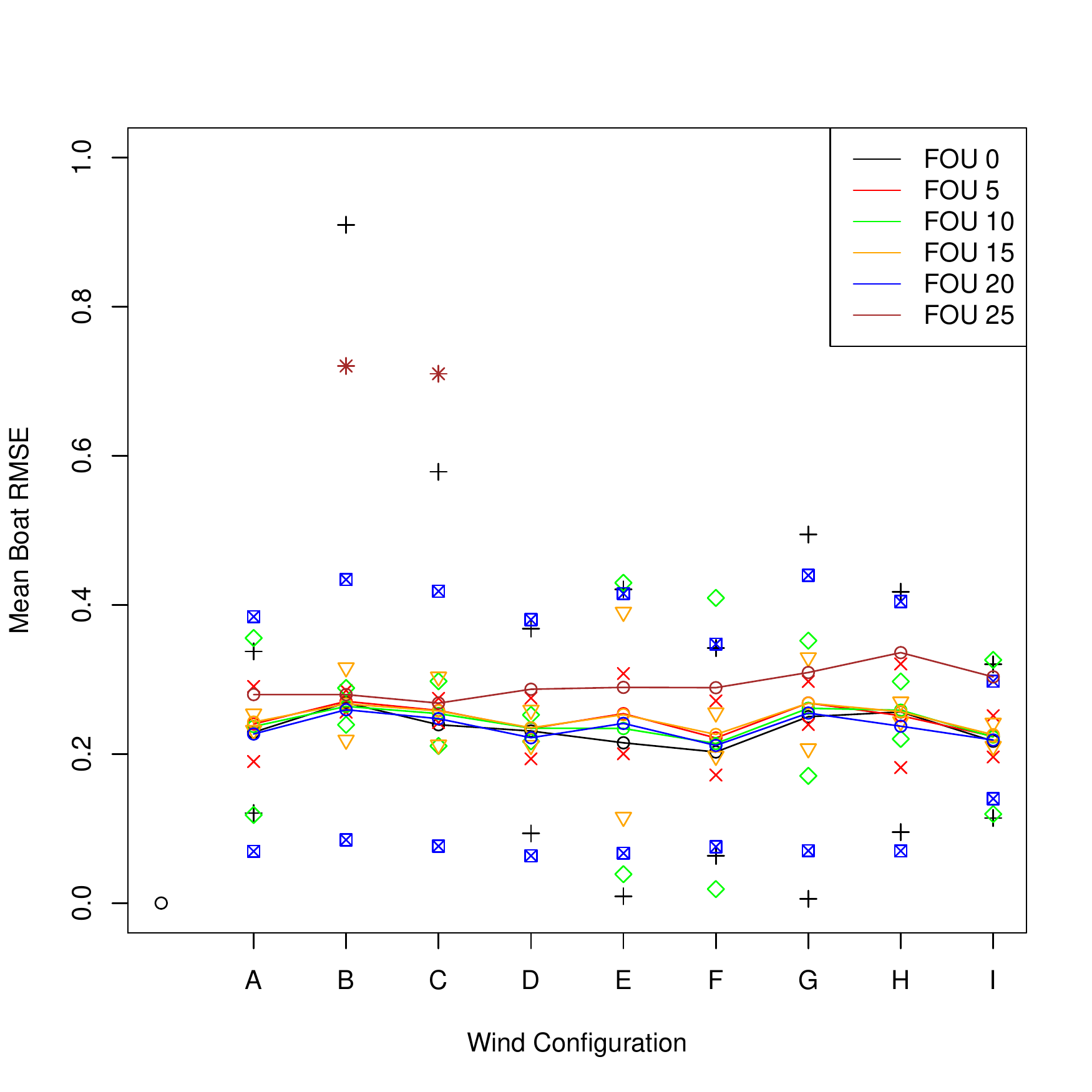}}
\subfigure[RMSE for 100 Turn]{\label{fig:RMSE_100}\includegraphics[scale=0.3]{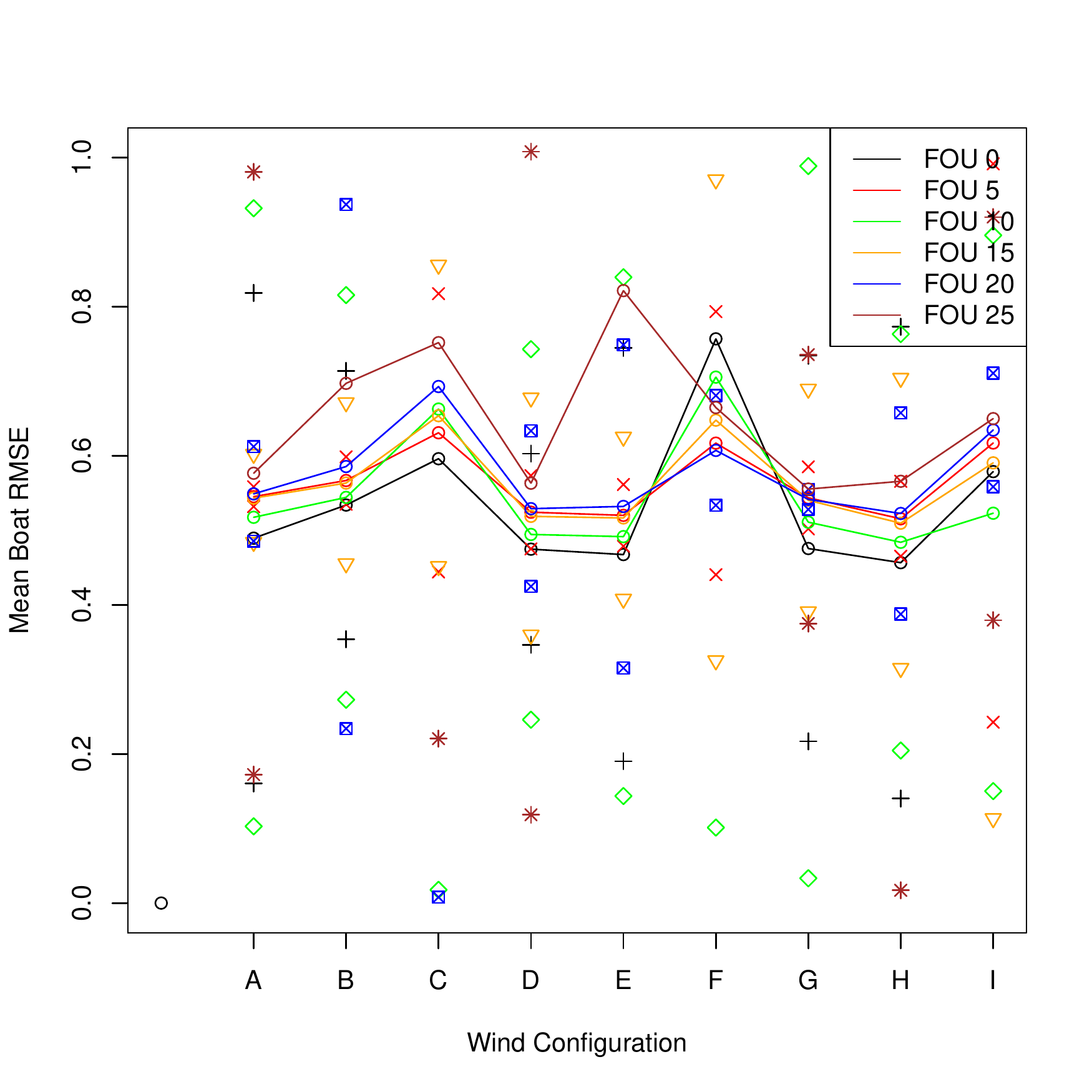}}\\

\subfigure[Uncertainty Measure for 25 Turn]{\label{fig:UM_25}\includegraphics[scale=0.3]{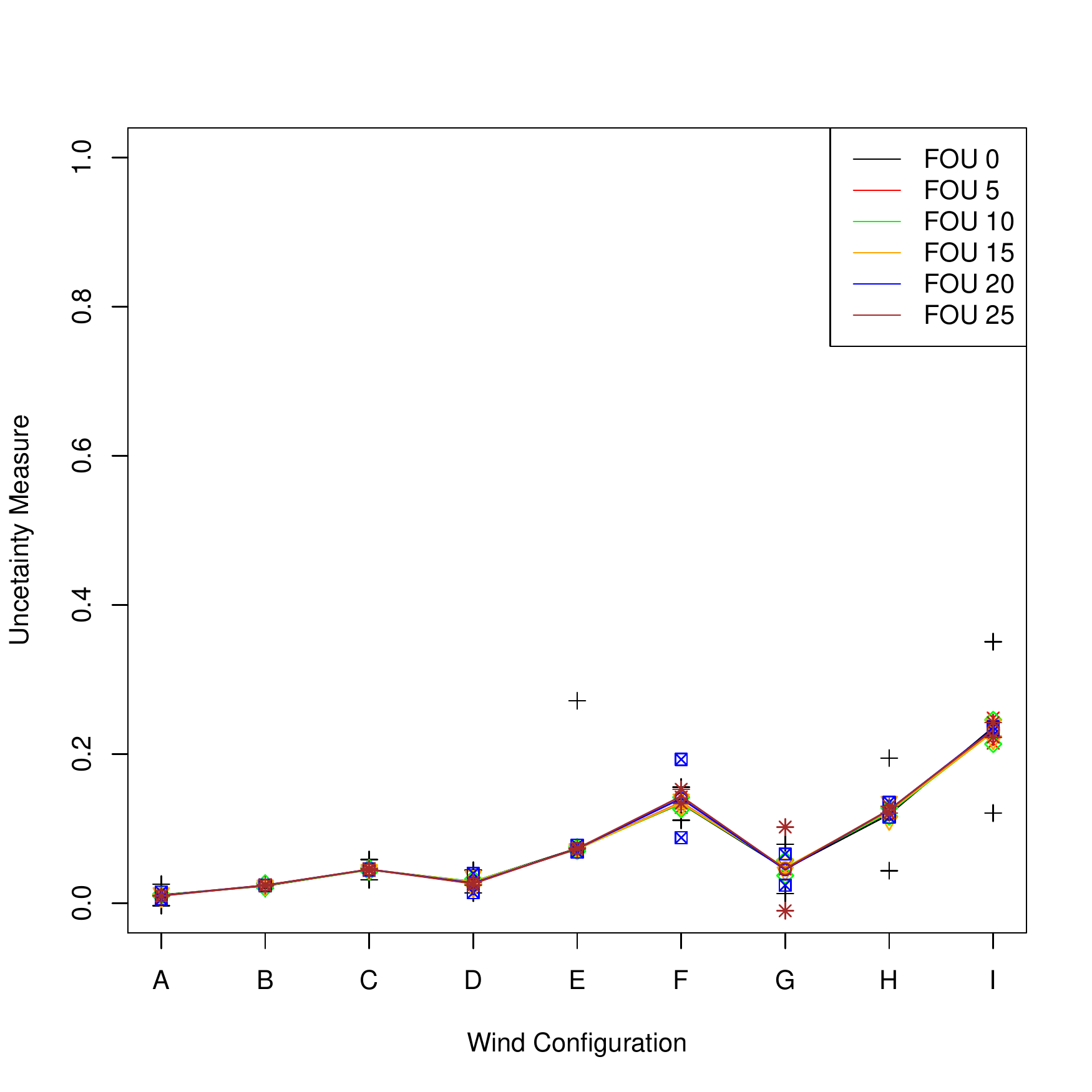}}
\subfigure[Uncertainty Measure for 50 Turn]{\label{fig:UM_50}\includegraphics[scale=0.3]{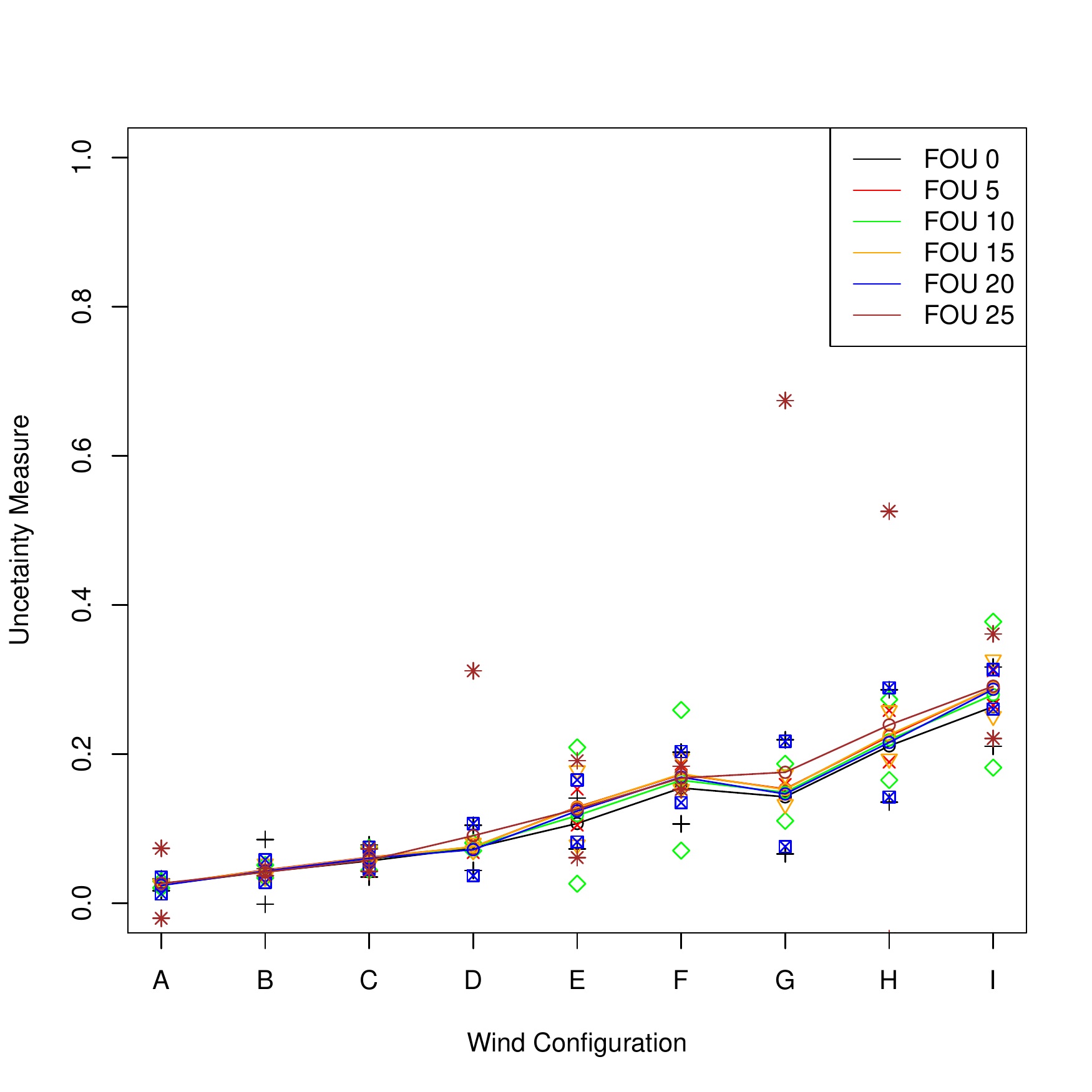}}
\subfigure[Uncertainty Measure for 100 Turn]{\label{fig:UM_100}\includegraphics[scale=0.3]{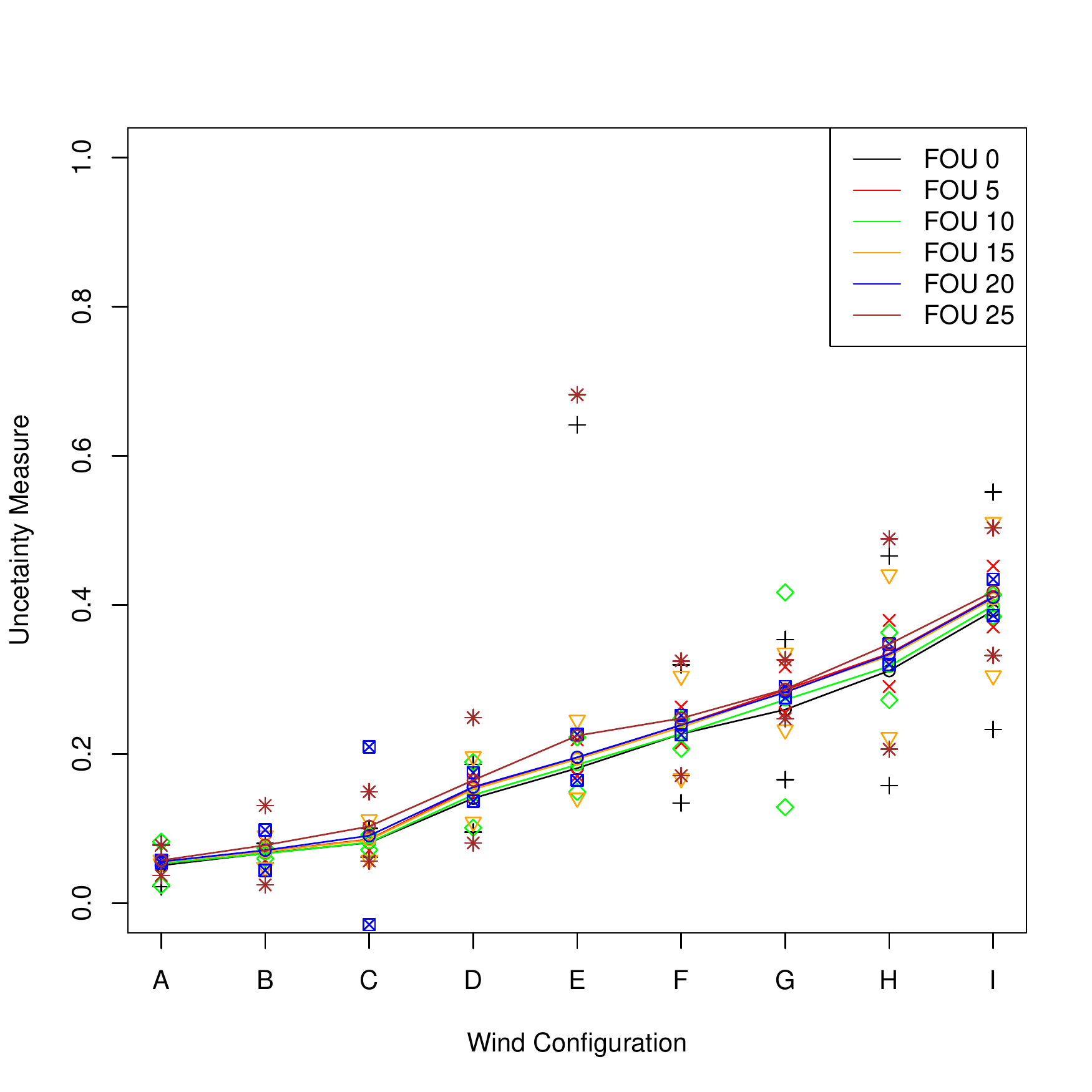}}\\

\subfigure[Relative Performance for 25 Turn]{\label{fig:RP_25}\includegraphics[scale=0.30]{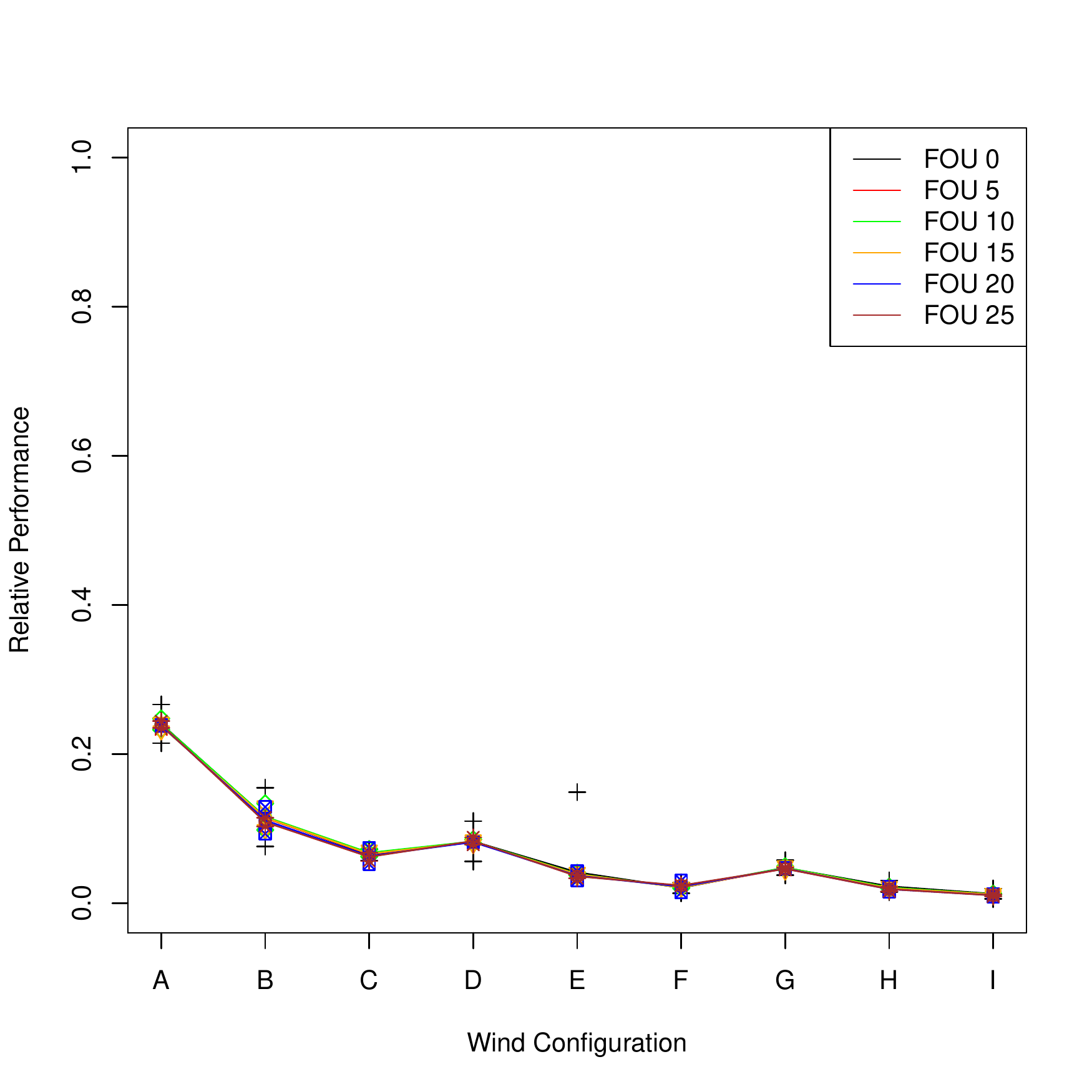}}
\subfigure[Relative Performance for 50 Turn]{\label{fig:RP_50}\includegraphics[scale=0.3]{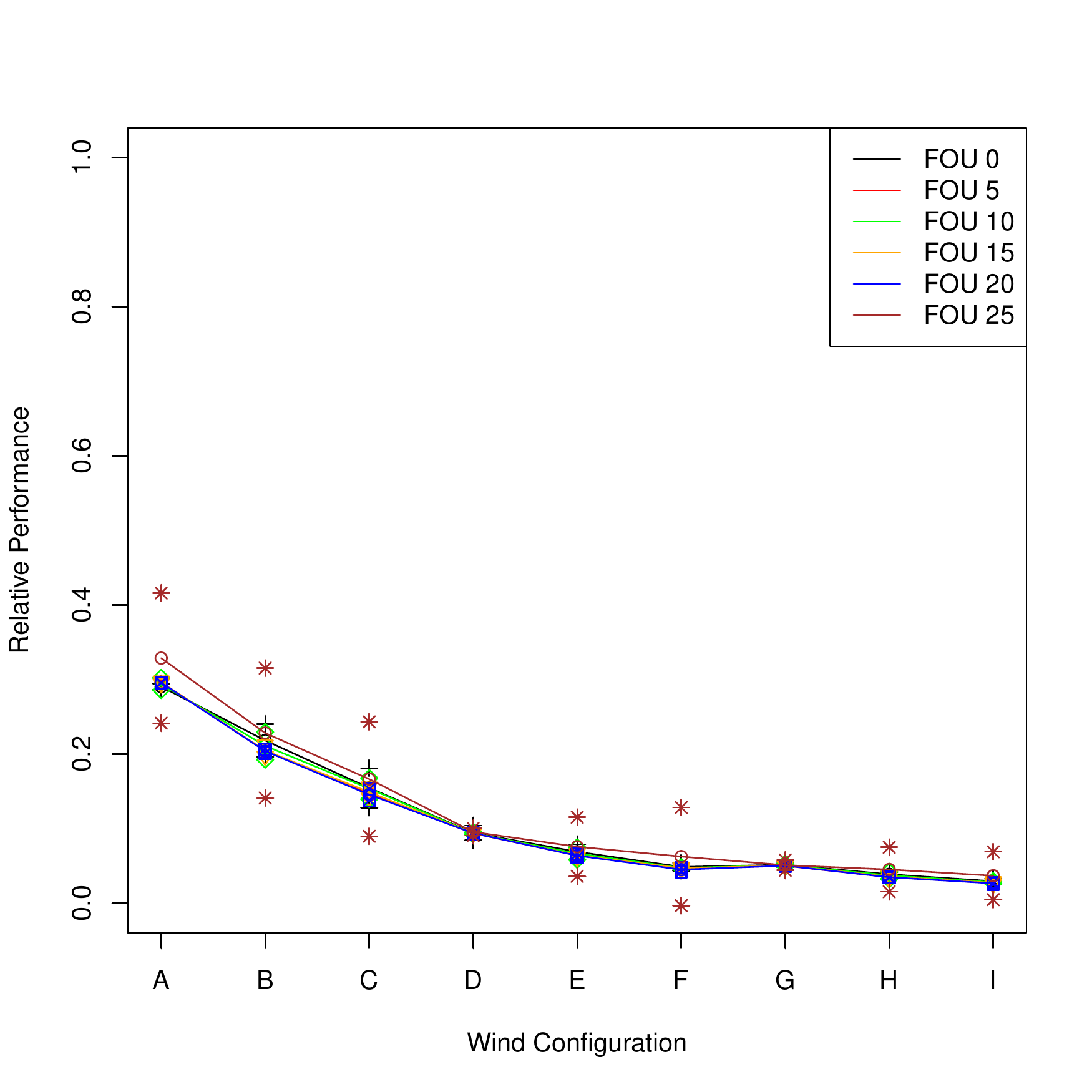}}
\subfigure[Relative Performance for 100 Turn]{\label{fig:RP_100}\includegraphics[scale=0.3]{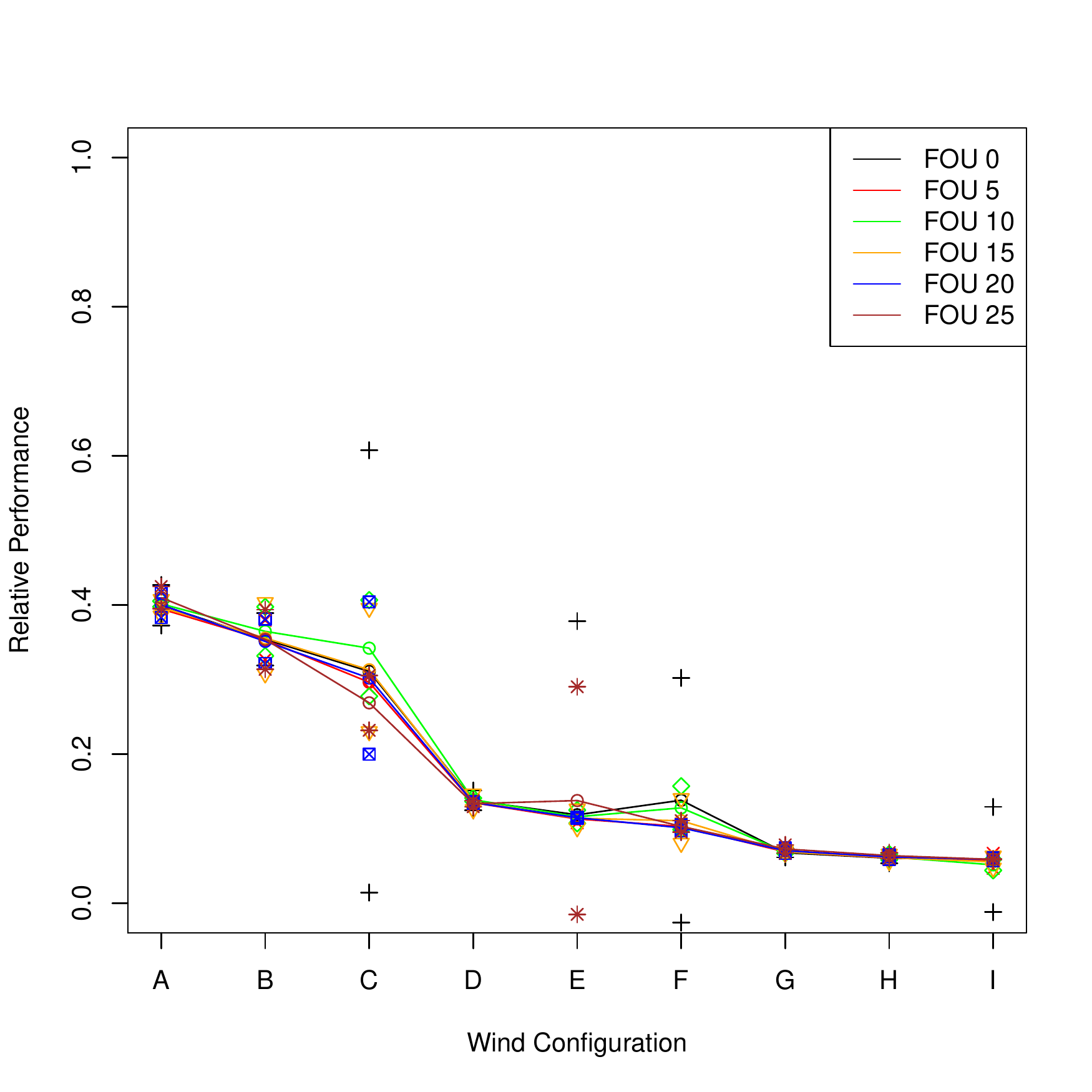}}
\caption{Absolute Performance (RMSE) (top) and uncertainty measure (middle) used calculate Relative Performance (Bottom).  Values are scaled betwen 0 and 1.}
\label{fig:RMSEandRP}
\end{figure*}

	Figures in \ref{fig:RMSEandRP} shows, row by row: Absolute Performance (RMSE), followed by Uncertainty Measure and finally Relative Performance which is composed of the top two (as well as the Base difficulty which is not shown here) inputs.  Each graph shows the specified output as the wind configuration changes from `A' to `I' with each line representing a different footprint of uncertainty configuration in the fuzzy controller under consideration.  Several things can be observed:  Firstly in each set of figures, the vertical positioning of each graph increases.  This indicates that overall performance, as measured by both Relative Performance and RMSE gets worse as the vertical movement is increased.  We believe this is a logical conclusion that can be made as a greater vertical movement is more difficult to complete.  Secondly we can observe that in the case of Relative Performance every line is very close together indicating that in fact the FOU does not seem to affect the relative performance as much as the Uncertainty present in the environment in these experiments.
	
	Tables \ref{Tab:FOU0Metrics} \ref{Tab:FOU10Metrics} \ref{Tab:FOU20Metrics} show the three input variables and resulting Relative Performance values for a single course in which 50m of vertical movement is required to complete.  Each Table enumerates the values for different FOU sizes, 0, 10 and 20 respectively and it can be observed that the uncertainty values remain relatively constant, an expected result as in each experiment the wind conditions are repeated.  The fact that the values do not change in the different tables give further evidence that FOU does not in fact affect the performance of type-2 controllers in these experiments.

\section{Discussion}
\label{sec:Discussion}

We have shown that the technique described in this paper gives different results to the more traditional method of comparing runs of simply using the RMSE as has been done in previous investigations \cite{Benatar2011}.  The resultant relative performance calculations show that the FOUs used in the experiment does not cause a significant difference in the conditions tested.  It is possible that the uncertainty levels used in this investigation are not sufficient for this to occur will be considered further in future work.  However the use of multiple different FOU sizes increases our confidence that this method is robust over different levels of FOU size.

The use of RMSE as shown in the graphs as the only performance metric is shown to have some issues, firstly Figure \ref{fig:RMSE_25} shows a flat line, which indicates that the RMSE does not change as the wind configuration changes in this experimental set-up, which is counter intuitive as logically under more difficult conditions performance should change in some way, specifically in the majority of cases decrease.  As the course is made more difficult by increase the angle of the turn required, the spread of results increases and several peaks become apparent most significantly in Figure \ref{fig:RMSE_100}, specifically at wind configurations `E' and `F', this is an issue as these peaks are unanticipated and do not appear in the corresponding relative performance graph Figure \ref{fig:RP_100} and at the moment the reasons for this are not understood.
	
	The increasing spread of results shown in the RMSE graphs and discussed above does not occur in the relative performance graphs as shown in Figures \ref{fig:RP_25} to \ref{fig:RP_100}.  If the relative performance is considered to be accurate then the FOU does not effect the performance of a controller at a given environmental set-up.  Additional work would be required to show if this is really the case or if the type or magnitude of the environmental uncertainty used here is not sufficient to cause significant differentiation.
  
	As an initial attempt at developing a more sophisticated technique we believe that the idea and concepts involved in the calculation of relative performance holds merit.  As this is the first application in which it has been applied it is believe that there are still some shortcomings present which should be investigated in future work and are discussed below.
 
One of the large potential short comings with this technique is that if applied to real world applications, it could be difficult to identify and then measure each of the potential sources of environmental uncertainty.  Real life wind for example is a much more complicated physical process than the simulated process and measuring it is much more difficult, with behaviours such as gusting occurring and the fact that two accurate sensors located in different locations may give significantly different readings.

From the flow of data shown in Figure \ref{fig:MetricDesign} we have observed that at this point only the mean values are used from the ``Raw Metrics'' data source.  In future work it needs to be investigated if inclusion of the standard deviation of the ``raw metrics'' further improves the behaviour metric.  This will be done by defining an additional input to relative performance termed ``Performance Strength'' to create a symmetrical data flow from each data source, ``Uncertainty Sources'' and ``Raw Metrics''.   The use of average speed as an additional piece of data from the ``Raw Metrics'' will also be considered.

In this paper both of the sources of uncertainty, the wind direction and wind speed under study are given equal weighting and this may not be a true assumption.  It may be that an increasing standard deviation of the wind direction is considerably harder to cater for by the controller than the wind speed.  Further study is required in order to see if this is the case and it is possible that a scaling measure may be required in order to give one source greater influence than the other in this application.  This reasoning may also be applied to the base difficulty and absolute performance metrics, though in the latter case an additional input would be required --- another avenue for future investigation.

\subsection{Future Work}

The types of uncertainty studied in this paper are also not the only sources that are important in a the chosen application, especially when a real world environment is considered as instead of simulation.  The FLOATs platform can also be run on a physical robot, and in this case the effect of a number of additional sources of uncertainty must be considered.  For example the robot in question has a simple motor encoder based wind vane that is only able to give a wind direction to a single degree of accuracy and at a certain rate, which means the reading of the wind direction taken may be a considerable source of uncertainty.

Additionally, each of the inputs that are used in the final calculation should be the subject of its own in depth investigation in order to ensure that it is modelled in the most accurate and meaningful way possible.  This is especially true of the uncertainty measure, as uncertainty is such an important idea in the field of fuzzy logic and a greater understanding of its effect upon these sorts of systems is required in order to allow logical reasoning when considering such questions as the selection of the appropriate fuzzy variety in a given application.

\section{Acknowledgements}

	The authors would like to thank the School of Computer Science, University of Nottingham for their support in funding this paper.


\bibliographystyle {ieeetr}
\bibliography{Journal}



\end{document}